\pdfoutput=1
\documentclass[11pt]{article}
\usepackage{ACL2023}
\usepackage{times}
\usepackage{latexsym}
\usepackage[T1]{fontenc}
\usepackage{multirow}
\usepackage{amssymb}
\usepackage{graphicx} 
\usepackage{booktabs} 
\usepackage{amsmath} 
\usepackage[utf8]{inputenc}
\usepackage{microtype}
\usepackage{inconsolata}
\title{UniMC: A Unified Framework for Long-Term Memory Conversation via Relevance Representation Learning}

\author{Kang Zhao,  Wei Liu, Jian Luan, Minglei Gao, Li Qian, Hanlin Teng, Bin Wang\\
  XiaoMi AI Lab\\
  {\tt zhaok7878@gmail.com, liuwei40@xiaomi.com, luanjian78@hotmail.com}, \\{\tt \{gaominglei,qianli,tenghanlin1,wangbin11\}@xiaomi.com
}\\
  }

\begin{document}
\maketitle
\begin{abstract}
Open-domain long-term memory conversation can establish long-term intimacy with humans, and the key is the ability to understand and memorize long-term dialogue history information. Existing works integrate multiple models for modelling through a pipeline, which ignores the coupling between different stages. In this paper, we propose a \textbf{Uni}fied framework for Long-term \textbf{M}emory \textbf{C}onversations (UniMC), which increases the connection between different stages by learning relevance representation.
Specifically, we decompose the main task into three subtasks based on probability graphs: 1) conversation summarization, 2) memory retrieval, 3) memory-augmented generation.
Each subtask involves learning a representation for calculating the relevance between the query and memory, which is modelled by inserting a special token at the beginning of the decoder input.
The relevance representation learning strengthens the connection across subtasks through parameter sharing and joint training. 
Extensive experimental results show that the proposed method consistently improves over strong baselines and yields better dialogue consistency and engagingness.

\end{abstract}

\section{Introduction}
Open-domain dialogue systems aim to establish long-term connections with users \cite{huang2020challenges}.
Current dialogue models based on pre-training \cite{DBLP:journals/corr/abs-2001-09977,DBLP:conf/eacl/RollerDGJWLXOSB21,bao2021platoxl,gu2022eva2} perform well on conversational relevance and fluency, with the advantage of capturing domain common sense and background knowledge. However, these models can neither handle long-term conversational context nor establish long-term connections with users.

Many studies on persona-based dialogue \cite{li-etal-2016-persona,zhang-etal-2018-personalizing,ijcai2018-595,song2019exploiting} facilitate in-depth chat by assigning profile information. These works mainly focus on the inconsistency of chatbot persona and context.
Nevertheless, the above works can only consider the consistency of the chatbot's persona but ignore the memory and utilization of the user's persona. In addition, these methods lack the long-term persona ability \cite{xu2022long}. This ability requires understanding and remembering long-term dialogue history information. After interacting with a human, the chatbot can memorize and update the personas of the user and itself, which is used appropriately to make the response more attractive and maintain the consistency of the dialogues.

\begin{figure}[t]
	\centering
	\scalebox{0.6}{
		\includegraphics{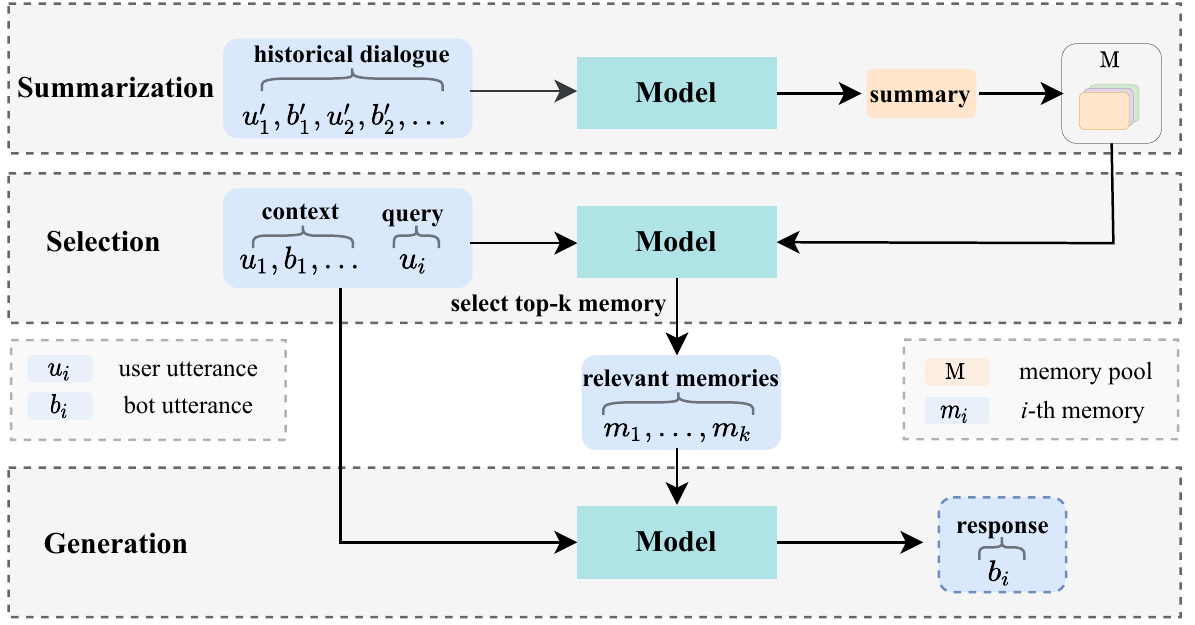}} 
	\caption{Our proposed framework (UniMC) for Open-Domain Long-Term Memory Conversation.}
	\label{fig1}
\end{figure}

Recent works have made addressed this issue. \citet{xu2022beyond} propose a long-term open-domain conversation task to advance research by constructing a dataset. It builds retrieval-augmented generative models and read-write memory-based models for modeling long-context. Nevertheless, its stored memories are not updated. \citet{xu2022long} constructed a multi-turn mutual persona dialogue dataset and proposed a long-term memory conversation task, in which they augmented dynamic summarization and memory updating.
However, it fails to summarize dialogue into persona but merely to classify sentences.
Moreover, these methods adopt the strategy of submodule execution, which a different model implements, and ignore the coupling between these modules.

To address the issues discussed above, we propose a unified modeling perspective that decomposes long-term memory conversation into related subtasks. These subtasks learn the interconnections across tasks through multi-task learning.

In addition, we propose a method to guide the execution of each subtask and strengthen the connection across them by learning relevance representation. As shown in Figure \ref{fig1}, we decompose the main task into three stages from a unified perspective based on probability graphs:
1) Summarization: the model inputs the context and outputs the memory that needs to be summarized.
2) Selection: the model retrieves the relevant memory by matching the context with the summary in the memory pool.
3) Generation: the model generates a response through the retrieved memory and context.
We add a special token for model's decoder input, which corresponds to the output representation as relevance representation. This represents the relevance between the query and memory pool proxy, which guides subsequent decoding or classification.

In summary, the main contributions of this paper can be summarized as follows:
\begin{itemize}
\item We propose a unified framework with multi-task learning to compact historical dialogue into memory, judge the relevance between the present query and the memory, and use the relevant memory to generate a response for the present query.
\item We enhance the connection across subtasks by using the same relevance judgment, which can explicitly guide model outputs.
\item Extensive experimental results show that the proposed method outperforms strong baselines and yields better dialogue consistency and engagingness.
\end{itemize}

\section{Related Work}

\subsection{Persona-based dialogue}
Persona-based dialogue generates consistent and attractive responses that utilize profile information. The critical challenge is making the chatbot exhibit a consistent personality to gain users’ trust and their long-term confidence \cite{huang2020challenges}.
Early work encodes personas into distributed embeddings \cite{li-etal-2016-persona}, using latent variables without explicit profile descriptions, which makes generating engaging utterances challenging.
In recent years,
\citet{zhang2018personalizing} propose a dialogue generation task conditioned on explicit profile information and presented the PersonaChat dataset, which made chit-chat more engaging.
\citet{yavuz-etal-2019-deepcopy} enables the model to generate responses tailored to persona descriptions by extending the pointer generator network \cite{see2017get}, which allows the decoder to engage and replicate external knowledge hierarchically.
\citet{song2019exploiting} propose a memory-augmented architecture to exploit persona information. It combines a conditional variational autoencoder model to address the one-to-many problem.
\citet{madotto2019personalizing} regard persona-based dialogue learning as a meta-learning paradigm that can generate personalized responses using only a few dialogue samples collected from the same user.
\citet{song2021bob} decompose persona-based dialogue generation into response generation and consistency understanding tasks, augmenting coherence constraints with additional natural language inference data.
However, these persona-based dialogue methods' personas are static and not summarized and memorized over the dialogue. 
This paper proposes a unified modelling framework to support the model's ability to update and memorize persona.

\subsection{Pre-trained Dialog Models}
Pre-trained language models (PLM) perform very well on many tasks \cite{radford2018improving,devlin-etal-2019-bert,raffel2020exploring,lewis2020bart,shao2021cpt},
which proves that fine-tuning PLM can yield better performance than training from scratch.
Due to the discrepancy between the dialogue corpus and the general document, much work on pre-trained dialog models has recently emerged \cite{DBLP:journals/corr/abs-2001-09977,DBLP:conf/eacl/RollerDGJWLXOSB21,zhou2021eva,gu2022eva2}.
\citet{zhang2020dialogpt} proposed DialoGPT to overcome bland and uninformative response generation through mutual information maximization training.
\cite{bao2020plato} proposed PLATO, a pre-training dialog model based on discrete latent variables, to solve the one-to-many problem. And then proposed to use curriculum learning \cite{bao2021plato} and larger-scale pre-training model \cite{bao2021platoxl} to improve the model's performance.
\cite{chen2022dialogved} introduces continuous latent variables in an enhanced transformer pre-training framework to increase the correlation and diversity of responses.
Recently, some works \cite{thoppilan2022lamda} have explored models with larger parameter scales and also tried to incorporate more external knowledge to enhance the dialogue.
However, most of these models do not possess long-term conversation capacity.
In this paper, we increase the long-term ability to interact with users based on pre-training models.

\section{Methodology}
\subsection{Task Decomposition}
Following \citet{xu2022long}, we first formalize the long-term memory conversation task definition.
Given a dialog context: $c=\{u_{1},b_{1},u_{2},b_{2},\dots,u_{t-1},b_{t-1}\}$ and a query $q=u_{t}$, where $u$ and $b$ represent the user and the chatbot, and $c$ may contain many sessions.
Long-term memory conversation aims to predict response $b_{t}$ based on long-session dialogue, with the formula $P(b_{t}|x)$, where $x=[c;q]$.

A complete long-term memory conversation model should be able to find relevant memories, generate reasonable responses based on relevant memories, and make timely memory updates by summarization. 
To model long-term chat between the user and chatbot, we decompose the task into three stages by Bayes' rule:
\begin{equation}
\begin{aligned}
P(b_{t}|x) \propto P(m|x_1) P(z|x_2, m')P(b_{t}|x_2,m',z),
\end{aligned}
\label{111}
\end{equation}
where $x=[x_1;x_2]$ is divided into historical sessions $x_1$ and current session $x_2$, $z$ represents the relevance between memory and query. $m'\in \mathrm{M}_{u}\cup \mathrm{M}_{b}$ represents query-relevant memories. 
Each speaker has a corresponding memory pool $\mathrm{M}$, consisting of a series of persona descriptions $m$. These persona descriptions are summaries of the speaker's historical sessions.
We define $\mathrm{M}_{u}=\{m^u_1, m^u_2, \dots, m^u_m\}$ and $\mathrm{M}_{b}=\{m^b_1, m^b_2, \dots, m^b_n\}$ to represent the persona memory pool of user and chatbot, respectively.
We expect $z$ also to guide the model to generate responses consistent with memory.

We interpret the first term in Eq.\ref{111} as a conversation summarization task, the second as a memory retrieval task, and the last as a memory-augmented generation task. Its corresponds to the three stages of summarization, selection, and generation. The intuition behind the decomposition is that for a complete dialogue modeling between the user and the chatbot, $x$ may contain many rounds of conversation or multiple sessions. For long-term conversational context, the model needs to summarize the dialogues from the current session, recall (retrieve) them appropriately, and reply based on the retrieved content.

\begin{figure}[t]
	\centering
	\scalebox{0.76}{
		\includegraphics{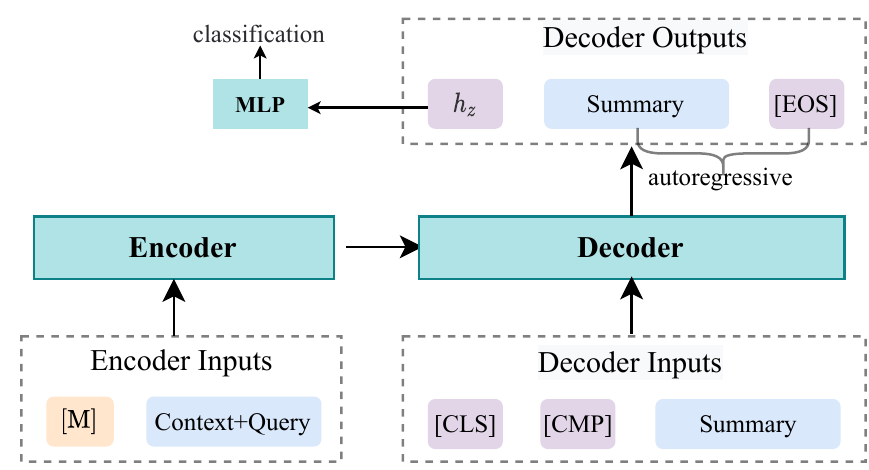}} 
	\caption{UniMC for the conversation summarization task, the token [M] is memory pool proxy token at the encoder. The first token [CLS] at the decoder is used to introduce relevance representation, and the second token [CMP] is used to generate the summary.}
	\label{write_fig}
\end{figure}

\subsection{Framework}
We model the long-term memory conversation task based on the Transformer \cite{NIPS2017_3f5ee243}, where the encoder encodes the context and memory, and the decoder is used for three stages: Conversation Summarization, Memory Retrieval and Memory-Augmented Generation. The three stages are unified in form and share the same model parameters.

\paragraph{Encoder}
Given the context $c$, we encode the context through the Transformer Encoder, where the role token is inserted into each utterance to distinguish the user and chatbot. 
Similarly, the input of persona memory is also distinguished by inserting role tokens, and [M] is inserted at each starting position to represent the memory input. The corresponding output $h^{[M]}$ of [M] at the encoding end is called the memory pool proxy. When only [M] is encoded, it represents a pattern of all memories. When specific memories is encoded together, it represents the specific pattern of these memories.

We use Fusion-in-Decoder (\textbf{FiD}) \cite{izacard2020leveraging} for memory integration, and conduct ablation for the Fusion-in-Encoder (\textbf{FiE}).
The latter is to concatenate the context and memory and feed it into the encoder, which belongs to the concatenation of text.

\paragraph{Decoder}
According to the task decomposition, we insert a special token [CLS] into the starting position of each subtask to encode the relevance representation. In order to distinguish different generation tasks, we insert special tokens [CMP] or [GNR] to represent the decoding of summarization or response generation.

\subsubsection{Conversation Summarization}
\begin{figure}[t]
	\centering
	\scalebox{0.76}{
		\includegraphics{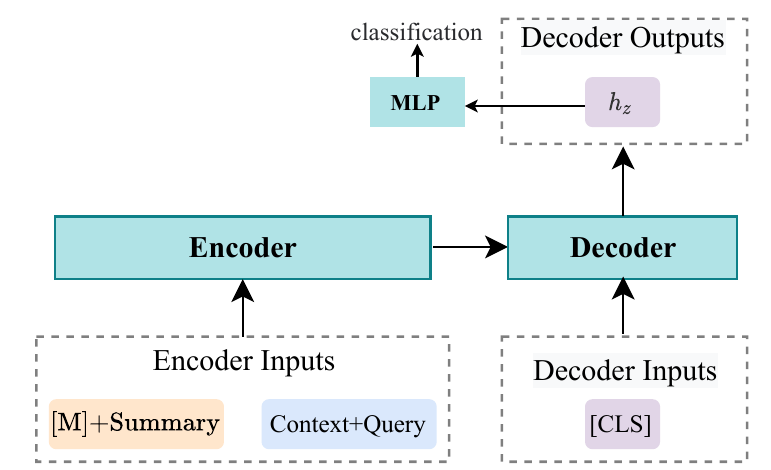}}
	\caption{UniMC for the memory retrieval task, which the decoder input only contain one token [CLS].}
	\label{read_fig}
\end{figure}

Acquiring memory from context can be viewed as a conversation summarization. Given the context $c=\{u_{1},b_{1},\dots,u_{t-1},b_{t-1}\}$ and query $u_{t}$, the model output the persona memory $m$. In Eq.\ref{111}, we mentioned that the relevance representation represents the semantic association of the query and memory. As shown in Figure \ref{write_fig}, we further introduce the memory pool proxy token [M]:
\begin{equation}
\begin{aligned}
P(m|x) = P(z|x) P(m |x, z),
\end{aligned}
\label{222}
\end{equation}
where $z$ contains the relevance between memory pool proxy and query, $x=\{ [\mathrm{M}]; u_{1},b_{1},\dots,u_{t-1},b_{t-1},u_{t}\}$ is model input. 

The memory pool proxy token is viewed as an abstract summary that represents the abstract \textit{pattern} of all persona memories. We compute $p(z|x, \mathrm{[M]}, w_{0}) = softmax(\mathrm{MLP}(h_{z}))$ as a binary classification task by decoding the relevance representation $h_{z}$ at the first start token $w_{0} = \mathrm{[CLS]}$ of the decoder. If the current query $u_t$ is related to the \textit{pattern}, then the query and context can be compacted into a summary, in which the relevance label $y_{cs}$ is 1; otherwise, it is 0.

We start decoding the sequence at the second token $P(m|x,z) =  {\textstyle \prod_{t=1}^{|m|}} p(w_{t}|x,z,w_{<t})$, where $w_{0}=[\mathrm{CMP}]$ is interpreted as the task identifier.
Finally, we minimize the negative log-likelihood loss for conversation summarization:
\begin{equation}
\begin{aligned}
\mathcal{L}_{cs}=& -\mathrm{log}P(m|x).
\end{aligned}
\label{333}
\end{equation}

\subsubsection{Memory Retrieval}

In Eq. \ref{111}, the memory retrieval can be modeled as a sentence pair classification task, which judge the relevance between the current query and the memory. As suggested by \cite{xu2022long}, the memory in each training sample can be divided into positive persona memory and negative persona memory.

The positive memory is defined as the persona used in the current user's utterances and the bot's responses (both bot persona and user persona observed by the bot). In contrast, the negative memory is the remaining persona in the current session. As shown in Figure \ref{read_fig}, given a context $c$, query $u_{t}$ and sample a summary $m$ from the memory pool $\mathrm{M}$, we learn the relevance representation at the first starting token of the decoder:
\begin{equation}
\begin{aligned}
\mathcal{L}_{mr}=& -\mathrm{log}P(z|x,m).
\end{aligned}
\label{444}
\end{equation}
where $P(z|x,m)=P(z|x,m,w_{0})$, $w_{0}=\mathrm{[CLS]}$ is the start token of the decoder input. 
Similar to the relevance judgment in conversation summarization, the label is determined based on whether the persona $m$ is positive or negative.
Note that $m$ represents just one persona, randomly sampled from both positive and negative personas, and the token [M] is inserted at the starting position of each $m$.

\subsubsection{Memory-Augmented Generation}

\begin{figure}[t]
	\centering
	\scalebox{0.76}{
		\includegraphics{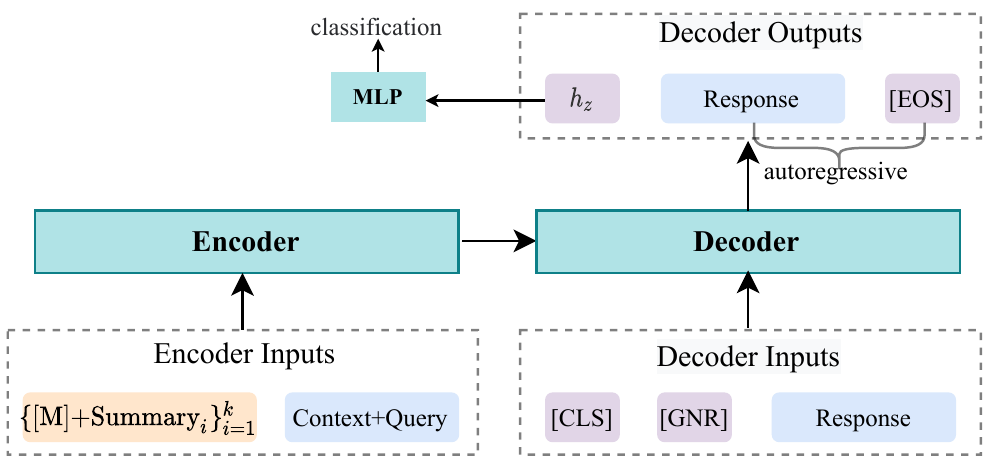}} 
	\caption{UniMC for the memory-augmented generation task, the first token [CLS] at the decoder input for modelling relevance representation, and the second token [GNR] generates the response.}
	\label{gen_fig}
\end{figure}
In the memory-augmented generation, we use the retrieved memory as external information to generate a more realistic reply. Similar to what is mentioned before, relevance representation also guides response generation. However, in the actual situation, not all of the memories retrieved from the large-scale memory pool may be relevant to the current conversation round. To address this issue, we add some noise to the retrieved memories by randomly introducing some negative memories. This allows the relevance representation to decide whether to use these memories, which can alleviate the problem of overrecall. The task is then constrained with a classification:
\begin{equation}
\begin{aligned}
P(b_{t}|x,m) = P(z|x,m) P(b_{t} |x, m, z),
\end{aligned}
\label{555}
\end{equation}
unlike the retrieval task, there may be multiple summaries $m$ introduced here.
If $m$ is all composed of negative memories, the relevance label is 0; otherwise, it is 1. The $P(b_{t} |x, m, z) =  {\textstyle \prod_{t=1}^{|b_{t}|}} p(w_{t}|x,m,z,w_{<t})$ is then trained using a negative log-likelihood loss, and $w_{0}=[\mathrm{GNR}]$ is interpreted as the task identifier. The complete loss can be written as: $\mathcal{L}_{mag}=-\mathrm{log} P(b_{t}|x,m)$, mag = "memory augmented generation".

We train these three subtasks simultaneously for the entire long-term memory conversation. 
Furthermore, each subtask contains the judgment between query and summary, which strengthens the connection across different tasks during training.
Our final training loss is:
\begin{equation}
\begin{aligned}
\mathcal{L}=&-logP(b_{t}|c) 
=\mathcal{L}_{cs} + \mathcal{L}_{mr} + \mathcal{L}_{mag}.
\end{aligned}
\label{666}
\end{equation}

\subsection{Inference}
We store all memory into a memory pool $\mathrm{M}=\mathrm{M}_{u} \cup \mathrm{M}_{b}$ and treat it as the long-term memory of the model. In the inference phase, the model first \textbf{Retrieval} the memory pool: performs retrieval tasks based on the memory pool and the given context and query, then takes top-k as the relevant memory according to logits. Then, \textbf{Generation}: performs the memory-augmented generation based on the context and associated memory. Finally, \textbf{Summarization}: perform the conversation summarization according to the context and query, extract the memory of the user and chatbot, respectively. The encoder encodes each memory, and the representation corresponding to the first start token [M] is taken as the memory vector and then written according to the cosine similarity:
\begin{equation}
\begin{aligned}
s =cos(h_{m_{i}}^{[M]}, h_{m_{j}}^{[M]}),
\end{aligned}
\label{666}
\end{equation}
where $m_{i} \in \mathrm{M}$ and $m_{j}$ is the memory that the model summarizes from the context.
When $s>\lambda$, we replace $m_{i}$ in $\mathrm{M}$ by $m_{j}$, otherwise store $m_{i}$ directly into $m_{j}$, where $\lambda$ is a duplicate check threshold.

\paragraph{Explicitly Guided Decoding}
The decoded tokens are implicitly influenced by relevance representation for the conversation summarization and memory-augmented generation subtasks. However, relevance representation can represent memory-related semantic information of the decoded text.

To discover the role of relevance representation, we propose an explicitly guided decoding by relevance representation method (EG), which improves the generation performance by explicitly guiding the decoding process:
\begin{equation}
\begin{aligned}
w_{t}=
\underset{w \in V^{(k)}}{\arg \max }\{
p\left(w|w_{<t}\right)+\alpha \cdot p(w| h_{z})\},
\end{aligned}
\label{EG}
\end{equation}
where $p\left(w|w_{<t}\right)$ is the probability that the model generates the next token.

\section{Experiments}

\subsection{Datasets}
Our experiments are conducted on DuLeMon \cite{xu2022long}, the largest dataset of multi-turn Chinese mutual persona chats currently available. DuLeMon consists of two parts, DuLeMon-SELF and DuLeMon-BOTH. In DuLeMon-SELF, the bot only knows its persona, while in DuLeMon-BOTH, the bot also knows part of the user's persona.

\subsection{Evaluation Metrics}
\paragraph{Automatic evaluation Metrics}
The BF1 and Rouge \cite{lin-2004-rouge} are used to evaluate the conversation summarization. 
The BF1(\%) denotes the harmonic mean of the precision and recall. It evaluates whether the model recognizes a persona in the context that needs to be summarized.
Following \cite{xu2022long}, we use AUC and Recall@k to evaluate the ability of persona memory retrieval. The PPL, BLEU \cite{DBLP:conf/acl/PapineniRWZ02}, F1(\%), DISTINCT-1/2 \cite{DBLP:journals/corr/abs-2003-07568}, and the
model-based BERTScore \cite{zhang2019bertscore} are used to evaluate generated and human-annotated reference responses.
\paragraph{Human Evaluation Metrics}
In human evaluation, we use three utterance-level measures: coherence, consistency, and engagingness. The metrics is similar to previous research \cite{xu2022long}, but we score coherence and consistency on a scale from 0 to 1. For engagingness, we have a different definition. This better reflects the plausibility of the generated responses. Four crowdsourced workers evaluate each response, and a higher score indicates better quality.

Our discussion for these three metrics are shown in Appendix \ref{sec:scoredetail}.

\begin{table}[t]
\centering
\small
\begin{tabular}{lccccc}
\toprule
Model & Backbone  & BF1  & Rouge-1/2/L \\ 
\midrule \addlinespace[0.1cm]
UniMC & CPT-base  & 86.32 & 0.675/ 0.480/ 0.663   \\
UniMC & CPT-large & 87.40 & 0.787/ 0.658/ 0.782   \\ 
\bottomrule
\end{tabular}
\caption{Experimental results of different summarization methods.}
\label{sumtask}
\end{table}

\subsection{Baselines}
We evaluate UniMC and several strong baselines on DuLeMon for comparison: 

\begin{itemize}
\item EVA 2.0 \cite{gu2022eva2}: EVA2.0 is a state-of-the-art Chinese dialogue generation model.
\item EVA2.0-FT: The EVA2.0 model is fine-tuned on the DuLeMon dataset.
\item CPT-FT: The CPT model \cite{shao2021cpt} fine-tuned on DuLeMon dataset. CPT is a Chinese pre-trained unbalanced transformer model.
\item UniMC: UniMC is our proposed relevance representation-based unified modelling framework for long-term memory conversation.
\end{itemize}

\subsection{Training Details}
The large-scale model (CPT-large and EVA2.0-xlarge) is first pre-trained during training, where the pre-trained corpus is the DuLeMon-SELF training set. Then, the model is finetuned, and the finetuned corpus is the DuLeMon-BOTH train set. The pre-trained checkpoints are chosen based on the perplexity of the DuLeMon-SELF validation set. More details of the model are shown in Appendix \ref{sec:detail}.

\subsection{Results and Discussion}

\subsubsection{Results of conversation summarization}
We evaluate the models on the test set to measure the performance of different models on conversation summarization. 
We use the user persona (unseen) in the test set as a positive sample of the need to summarize and other dialogues that do not need to summarize memory as negative cases. The samples that need to be summarized and those that are not required are 183 and 162, respectively.

As shown in table \ref{sumtask},
we can see that the BF1 score exceeds 87.4\%, which shows that our method effectively identifies whether the dialogue can is compacted into a summary. Other metrics are to evaluate how well the summary matches the reference persona. It can be seen that the CPT-large model consistently outperforms the CPT-based model due to the larger amount of parameters. The experimental results show that the UniMC can effectively extract persona information from the dialogue.

\subsubsection{Results of Memory Retrieval}
In comparing different memory retrieval models, we added the CPM baseline \cite{xu2022long}, a rank model and reported the metrics on the test set. Table \ref{readd} shows results that comparison of different baselines on the retrieval task. We find that the UniMC consistently outperform CPM, which UniMC(CPT-large) outperforms CPM by 7\% and 10\% on AUC and Recall@5, indicating that our unified modeling approach demonstrates significant advantages.
The experimental results show that our method efficiently retrieves context-related persona.

\subsubsection{Results of Memory-Augmented Generation}
On the memory-augmented generation, we compare UniMC and direct fine-tuning with different baselines. Table \ref{main_auto} shows the comparison results of different models, in which the results of PLATO-FT are copied from the original paper \cite{xu2022long}. From these results, we can draw several conclusions:
\begin{table}[t]
\centering
\small
\begin{tabular}{lccc}
\toprule
Model & Backbone  & AUC  & Recall@5 \\ \midrule \addlinespace[0.1cm]
CPM\dag   & ERNIE     & 0.76 & 0.83     \\ \hline \addlinespace[0.1cm]
UniMC & CPT-base  & 0.80 & 0.91     \\
UniMC & CPT-large & \textbf{0.83} & \textbf{0.93}     \\ \bottomrule
\end{tabular}
\caption{Comparison of different memory retrieval models. The result CPM \dag copied from the original paper\cite{xu2022long}.}
\label{readd}
\end{table}

1) In the experimental results of different backbone fine-tuning, it can be seen that the performance of different backbones significantly differs. In the perplexity metric, this cannot be compared due to the differences in vocabulary and tokenization. However, in terms of other automatic evaluation indicators, the dialogue generation based on CPT fine-tuning is the best. It shows that the pre-trained model that considers understanding and generation may be better than other pre-trained dialog models in long-term memory conversation.

\begin{table*}[ht]
\centering
\small
\begin{tabular}{@{\extracolsep{4pt}}lcccccc}
\toprule
Model      & Backbone      & PPL   & BLUE-1/2    & DISTINT-1/2 & F1    & BERTScore \\ 
\midrule
\addlinespace[0.1cm]
PLATO-FT \dag  & PLATO-2   & 9.38 & 0.194/0.087 & \textbf{0.068}/\textbf{0.296} & 22.61 & -     \\
EVA2.0-FT  & EVA2.0-base   & 21.16 & 0.145/0.050 & 0.042/0.167 & 18.38 & 0.6215     \\
CPT-FT     & CPT-base      & 15.31 & 0.230/0.096 & 0.055/0.243 & 24.63 & 0.6394     \\
EVA2.0-FT  & EVA2.0-xlarge   & 13.95 & 0.179/0.088 & 0.044/0.170 & 18.69 & 0.6272     \\
CPT-FT     & CPT-large      & 12.39 & \textbf{0.243}/\textbf{0.102} & 0.057/0.249 & \textbf{25.81} & \textbf{0.6449}     \\
 \hline \hline \addlinespace[0.1cm]
UniMC & EVA2.0-xlarge & 13.61 & 0.174/0.064 & 0.065/0.259 & 20.27 & 0.6336 \\
UniMC & CPT-base      & 13.72 & 0.209/0.086 & \textbf{0.102}/\textbf{0.392} & 25.38 & 0.6399     \\ 
UniMC & CPT-large     & \underline{9.54}   & \textbf{0.217}/\textbf{0.090} & 0.084/0.339 & \textbf{26.14} & \textbf{0.6422}     \\ 
\bottomrule
\end{tabular}
\caption{Experimental results of the automatic evaluation for response generation and long-term memory conversation. The first five models are fine-tuned based on backbones without the capability of long-term memory conversation, so these methods cannot be directly compared with other long-term memory conversation models. We bold the best results in different scenarios, respectively. Perplexity (PPL) is not comparable between models due to the differences in vocabulary and tokenization. The result PLATO-FT \dag copied from the original paper\cite{xu2022long}.}
\label{main_auto}
\end{table*}

2) In the model comparison with large-scale parameters, UniMC based on CPT-large performs better than EVA2.0-xlarge and has fewer parameters than EVA2.0-xlarge. This result on automatic metrics further illustrates that non-dialog domain-specific pre-trained models seem to perform better on this task than pre-trained dialog models. This result is somewhat similar to the conclusion in \cite{zheng2021exploring} because we insert some special tokens on the decoder input to encode relevance representations and distinguish between different tasks.

The experimental results in Table \ref{main_auto} show that UniMC based on CPT-large performs better on most automatic metrics. Therefore, CPT-large-based UniMC is used for our subsequent human evaluation.

\subsection{Human Evaluation}
Automatic metrics are limited in evaluating open-domain dialogue tasks \cite{DBLP:conf/emnlp/LiuLSNCP16}. To further validate the model's performance, we conduct a self-chat evaluation. Self-chats are widely used in evaluating dialogue systems \cite{li2016deep,DBLP:conf/eacl/RollerDGJWLXOSB21,bao2021plato,xu2022long}, where the model plays the roles of both parties in the dialogue. Following \cite{xu2022long}, we use the proposed UniMC as a user simulator and ask all chatbots to chat for better control over variables. Crowdsourcing workers only evaluate the responses generated by the chatbot and not the user simulator. Details are as follows.

Each chatbot chats with the user simulator for ten episodes, each episode contains four long sessions, and each session contains 16 rounds. As described in \cite{xu2022long}, we do not impose any restrictions on a chat other than specifying session openings. Some conversation openings are pre-selected from the DuLeMon test set. These openings are used to start an interactive conversation and ask the two bots to perform a chat in a given context. Table \ref{human_eval} presents the result, from which we can draw several conclusions:

1) The unified modeling method can significantly improve the coherence and engagingness of the dialogue. Regarding dialogue coherence and engagingness, the model achieves scores of 0.796 and 0.824, which are substantially better than the baseline model EVA2.0. UniMC is 0.058 and 0.263 higher than CPT-FT on these two metrics, which indicates that our method can improve the model's dialogue performance and attract users to chat for multiple rounds. In addition, CPT-FT outperforms EVA2.0-FT on these two metrics, indicating that non-dialogue pre-trained models may be more effective for fine-tuning.

2) The unified modeling method can significantly improve dialogue consistency. In the evaluation, consistency needs to consider context consistency and persona consistency. EVA2.0 is more consistent than the fine-tuned EVA2.0. It is possible that direct fine-tuning makes it difficult for the model to maintain persona consistency in long-term dialogue. UniMC is more consistent than EVA2.0 and CPT-FT, showing that unified modeling of different subtasks can improve dialogue consistency.

\begin{table*}[ht]
\centering
\small
\begin{tabular}{lccc}
\toprule
Model     & Coherence & Consistency & Engagingness \\
\midrule \addlinespace[0.1cm]
EVA2.0    & 0.703         & 0.681           & 0.561            \\
EVA2.0-FT & 0.672         & 0.609           & 0.638            \\
CPT-FT    & 0.738         & 0.606           & 0.647            \\

UniMC  & \textbf{0.796}          & \textbf{0.740}           & \textbf{0.824}           \\
UniMC w/o EG     & 0.790 & 0.699 & 0.727    \\
\bottomrule
\end{tabular}

\caption{Comparison of human evaluation metric results for self-chat dialogues between UniMC and baselines. The parameter sizes of EVA 2.0 and CPT are xlarge and large, respectively. The `EG' represents that explicitly guided decoding strategies by relevance representation.}
\label{human_eval}
\end{table*}

\begin{table*}[ht]
\centering
\small
\begin{tabular}{@{\extracolsep{4pt}}ccccccccc}
\toprule
\centering
\multirow{2}{*}{Model} & \multicolumn{2}{c}{Generation} & \multicolumn{2}{c}{Retrieval} & \multicolumn{4}{c}{Summarization} \\  \cline{2-3} \cline{4-5} \cline{6-9} \addlinespace[0.1cm]
 & F1 & BERTScore & AUC & Recall@5 & BF1 & Rouge-1 & Rouge-2 & Rouge-L \\ \midrule
UniMC & \textbf{25.38} & \textbf{0.6399} & 0.80 & 0.91 & \textbf{86.32} & \textbf{0.675} & 0.480 & \textbf{0.663} \\ \hline \addlinespace[0.1cm]
\multicolumn{1}{r}{w/o EG} & 24.72 & 0.6370 & 0.80 & 0.91 & 84.49 & 0.667 & \textbf{0.491} & 0.655 \\
\multicolumn{1}{r}{w/o RR} & 24.55 & 0.6364 & 0.78 & 0.90 & 82.29 & 0.638 & 0.452 & 0.629 \\
FiD $\to$ FiE & 25.31 & 0.6370 & \textbf{0.82} & \textbf{0.92} & 80.53 & 0.668 & 0.481 & 0.656 \\
diff {[}CLS{]} & 24.15 & 0.6348 & 0.79 & 0.90 & 85.07 & 0.663 & 0.475 & 0.651 \\ \hline \hline \addlinespace[0.1cm]
diff decoder & 24.91 & 0.6380 & 0.82 & 0.92 & 82.92 & 0.680 & 0.482 & 0.667 \\ \bottomrule
\end{tabular}
\caption{Ablation studies result of UniMC. `EG' and `RR' denotes explicitly guided decoding and relevance representation, respectively. FiD$\to$ FiE denotes the method of integrating memory is converted from Fusion-in-Decoder to Fusion-in-Encoder. `diff [CLS]' denotes different tokens are used as the beginning of the decoder input. `diff decoder' denotes decoder parameters are not shared.}
\label{latent_abl}
\end{table*}

3) Explicitly guided decoding can significantly improve the performance of the model. UniMC can achieve scores of 0.796, 0.740, and 0.824 than without EG in coherence, consistency, and attractiveness. 
Experimental results show that introduced relevance representation may contain higher-level semantic information, which prove our proposed method's effectiveness and potential.
Moreover, we show some cases in Appendix \ref{sec:casestudy} to help illustrate the effectiveness of our model.

\section{Ablation Study}
\label{latent}
To further analyze the effectiveness of different components, we conduct ablation studies. The results of the ablation study are shown in Table \ref{latent_abl}, where our backbone adopts CPT-base. 

First, we remove EG when decoding, which the model's performance has decreased to varying degrees on the memory-augmented generation and dialogue summarization tasks. These experiments show that explicitly guided decoding can improve the quality of model generation, especially in conversation summarization, where it can significantly increase n-gram recall.
Second, we no longer model relevance representations and instead guide each task with a different token. Most metrics drop significantly on the three tasks, which illustrates the importance of relevance representation learning.
Next, we replace the memory integration method from FiD to FiE and observe that, while the retrieval task shows some improvement, other metrics have declined. Its shows the advantage of FiE in modelling long texts.
Finally, we use different start tokens to model the relevance representation for each task. Using the same token is more advantageous than this method because the same token modelling can further facilitate connections between tasks. Additionally, we conducted different tasks with different decoders, which increased the number of model parameters. The experimental results show that compared to the shared decoder method, there was not much performance improvement, which shows that our multi-task entire parameter sharing method is effective.

\section{Conclusions and Future Work}
This paper proposes a unified framework for long-term memory conversation, mainly by introducing relevance representation for multi-task learning.
Specifically, we decompose the main task into three subtasks and model each subtask with the same model. In addition, we insert a specific token into the decoder input to learn a relevance representations, which is used to guide the model's output. 
Our approach can better generate appropriate responses and enables a model to master retrieval and summarization simultaneously. Extensive experiments demonstrate the effectiveness of UniMC. Additional human evaluations show the advantages of UniMC.
UniMC has limitations in terms of interpretability and cannot provide reasons for why specific memories are used when generating responses, despite the fact that we use a sampling-based training strategy.
In the future, we will attempt to fuse neural-symbolic systems with pre-trained models, making the model logical and explainable in dialogue.

\bibliography{unimc}
\bibliographystyle{acl_natbib}

\appendix

\section{Scoring Criteria in Human Evaluation}
\label{sec:scoredetail}
The criteria used in human evaluation are provided in Table \ref{tab:criteria}, which each metric is at the utterance-level.
\begin{table*}[]
\centering
\small
\begin{tabular}{@{\extracolsep{4pt}}cl}
\hline
\multicolumn{1}{c|}{Score} & \multicolumn{1}{c}{Coherence} \\ \hline
\multicolumn{1}{c|}{0} & 
\begin{tabular}[c]{@{}l@{}} The response is not related with the context.\\ The response simply repeats the context.\\ The response has conflicts with the context.\\ There are any logic conflicts within the response.\end{tabular} 
\\ \hline
\multicolumn{1}{c|}{1} & The response is coherent with the context. \\ \hline
 &  \\ \hline
\multicolumn{1}{c|}{Score} & \multicolumn{1}{c}{Consistency} \\ \hline
\multicolumn{1}{c|}{0} & \begin{tabular}[c]{@{}l@{}}The response has conflicts with the context or memory.\\ There are logic conflicts within the response.\end{tabular} \\ \hline
\multicolumn{1}{c|}{1} & The response is consistent with context and with memory in dialogue history. \\ \hline
 &  \\ \hline
\multicolumn{1}{c|}{Score} & \multicolumn{1}{c}{Engagingness} \\ \hline
\multicolumn{1}{c|}{0} & \begin{tabular}[c]{@{}l@{}}I don't want to talk with this speaker.\\ The response  is kind of perfunctory or boring.\end{tabular} \\ \hline
\multicolumn{1}{c|}{1} & I would like to  talk with the speaker for each response in the  long-term conversation. \\ \hline
\multicolumn{1}{c|}{2} & The response is beyond expectations, not perfunctory, and more interesting. \\ \hline
\end{tabular}
\caption{Score details of metrics used in human evaluation.}
\label{tab:criteria}
\end{table*}

\section{Training Details}
\label{sec:detail}
We use a sampling strategy for memory retrieval and memory-augmented generation. In memory retrieval, a persona memory is randomly sampled from all seen memories for each utterance, and it is labeled as 1 or 0 depending on whether the sampled memory is positive or negative. The overall statistics of datasets are shown in Table \ref{duLeMon-statistics}. We scaled the memory retrieval data by five during training to balance the sample size of different subtasks. In addition, sampling training is also performed to simulate actual usage scenarios in the memory-augmented generation task. If there is no relevant persona for a round of dialogue, we sample the k personas as negative examples. The model can further judge whether the retrieved memory is valuable through such training.

For the memory of users and chatbots, we set the duplication threshold $\lambda$ as 0.9 and the number of candidates top-k to 3. We decode with beam search, where the beam size is set to 1 in the automatic evaluation and 4 in the self-chat evaluation. We fixed the random seed of the experiment as 2022. Following \cite{xu2022long}, we do not limit the memory capacity due to the sparsity of persona dialogue and the efficiency of our persona storage.

For all models, we set the maximum context length to 256 and used the default vocabulary of the pre-trained model. We optimize all models with the Adam \cite{DBLP:journals/corr/KingmaB14} optimizer with a learning rate of 5e-5. All models were trained using the Fairseq \cite{ott2019fairseq} toolkit and pre-trained models obtained from the transformers \cite{wolf-etal-2020-transformers} library. Experiments are conducted on NVIDIA V100 32GB and A100 40GB GPUs.

\begin{table}[]
\centering
\small
\begin{tabular}{lll}
\toprule
Category             & SELF & BOTH \\
\midrule
\# Dialogues                  & 24500       & 3001        \\
\# Utterances                 & 400472      & 48522       \\
Avg. \# turns                 & 16.3        & 16.2        \\
Avg. length of utterances     & 19.7        & 21.2        \\
Avg. \# bot persona           & 4.0         & 4.0         \\
Avg. \# user persona (seen)   & 0           & 4.4         \\
Avg. \# user persona (unseen) & 4.0         & 1.3        \\
\bottomrule
\end{tabular}
\caption{\label{duLeMon-statistics}Statistics of DuLeMon.}
\end{table}

\section{Case Study}
\label{sec:casestudy}
\begin{figure*}[t]
	\centering
	\scalebox{0.8}{
		\includegraphics{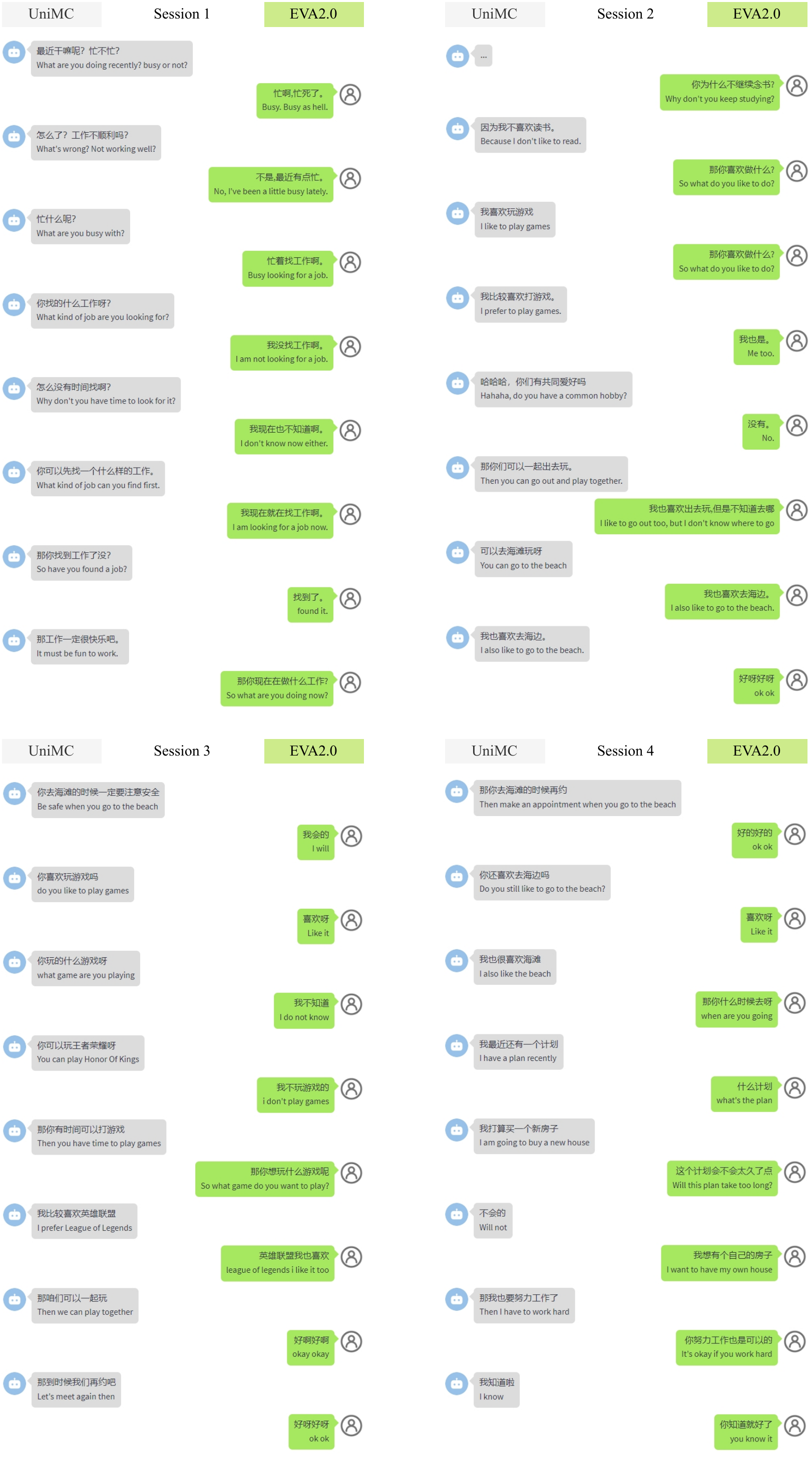}} 
	\caption{Example of one episode conversation between UniMC and EVA2.0.}
	\label{case}
\end{figure*}

To concretely demonstrate UniMC's ability to long-term memory conversation, we further provide a cherry-picked selected example of a single-episode dialogue (between UniMC and EVA 2.0) in Figure \ref{case}.

\end{document}